\pdfoutput=1
\documentclass{article}

\usepackage{arxiv}

\usepackage[utf8]{inputenc} 
\usepackage[T1]{fontenc}    
\usepackage{hyperref}       
\usepackage{url}            
\usepackage{booktabs}       
\usepackage{amsfonts}       
\usepackage{nicefrac}       
\usepackage{microtype}      
\usepackage{lipsum}
\usepackage{graphicx}
\usepackage[square,numbers]{natbib}
\usepackage{multirow}
\usepackage{float}
\usepackage{tabu}
\usepackage{caption}
\usepackage{amsmath}
\usepackage{algpseudocode}
\usepackage{hyperref}
\usepackage{authblk}

\usepackage{xcolor}
\usepackage{amssymb}
\usepackage{cancel}
\usepackage{booktabs} 
\usepackage{algorithm}
\usepackage{tcolorbox}

\title{Credible Plan-Driven RAG Method for Multi-Hop Question Answering}
\author[1]{Ningning Zhang}
\author[1]{Chi Zhang}
\author[1]{Zhizhong Tan}
\author[1]{Xingxing Yang}
\author[1]{Weiping Deng}
\author[1]{Wenyong Wang\thanks{corresponding author}
}
\affil[1]{Macau University of Science and Technology}

\begin{document}

\maketitle
\begin{abstract}
Retrieval-augmented generation (RAG) has demonstrated strong performance in single-hop question answering (QA) by integrating external knowledge into large language models (LLMs). However, its effectiveness remains limited in multi-hop QA, which demands both stable reasoning and factual consistency. Existing approaches often provide partial solutions, addressing either reasoning trajectory stability or factual verification, but rarely achieving both simultaneously. To bridge this gap, we propose PAR-RAG, a three-stage \textbf{P}lan-then-\textbf{A}ct-and-\textbf{R}eview framework inspired by the PDCA cycle. PAR-RAG incorporates semantic complexity as a unifying principle through three key components: (i) complexity-aware exemplar selection guides plan generation by aligning decomposition granularity with question difficulty, thereby stabilizing reasoning trajectories; (ii) execution follows a structured retrieve-then-read process; and (iii) dual verification identifies and corrects intermediate errors while dynamically adjusting verification strength based on question complexity—emphasizing accuracy for simple queries and multi-evidence consistency for complex ones. This cognitively inspired framework integrates theoretical grounding with practical robustness. Experiments across diverse benchmarks demonstrate that PAR-RAG consistently outperforms competitive baselines, while ablation studies confirm the complementary roles of complexity-aware planning and dual verification. Collectively, these results establish PAR-RAG as a robust and generalizable framework for reliable multi-hop reasoning.
\end{abstract}


\section{Introduction} \label{introduction}
\noindent  {R}{etrieval-augmented} generation (RAG) \cite{r:1} has emerged as a promising paradigm for enhancing large language models (LLMs) with external knowledge, driving progress in knowledge-intensive tasks such as open-domain QA and scientific discovery \cite{r:2, r:5, r:6}. However, in multi-hop question answering (MHQA) \cite{r:3, r:4}—where evidence must be retrieved from multiple sources and integrated through multi-step reasoning—existing RAG approaches remain inadequate. Two coupled challenges prevail: reasoning path deviation, in which intermediate steps diverge from the intended reasoning trajectory \cite{r:7, r:8}, and intermediate factual errors, often arising from hallucinations or irrelevant context \cite{r:9, r:10, r:11, r:12}. These errors tend to propagate across steps, compounding inconsistencies and undermining the reliability of final answers.

To address these challenges, prior studies have explored several complementary directions. Planning-guided methods impose explicit task decomposition to stabilize inference; self-reflection approaches introduce verification mechanisms to enhance factual alignment; structure-aware systems leverage graph-based or hierarchical representations to capture inter-document relations; and iterative reasoning frameworks refine intermediate outputs through repeated cycles \cite{r:15, r:16, r:17, r:21}. While each paradigm contributes partial improvements, none provides a unified solution. Reflection-based and reasoning-enhanced approaches improve factual accuracy but often destabilize global reasoning trajectories. In contrast, planning-guided methods reinforce trajectory stability but rely on weak or ad hoc verification. Finally, structure-aware systems emphasize data structure optimization yet lack both robust reasoning capabilities and effective mechanisms for verifying intermediate results.

More critically, most methods neglect the role of semantic complexity. Exemplar selection for plan generation is usually static, misaligning reasoning granularity with problem difficulty, while verification is often collapsed into a single accuracy score. Yet recent work on complexity-aware metrics and fine-grained evaluation \cite{r:63, r:61,r:64,r:65} shows that both reasoning performance and trustworthiness are shaped by semantic complexity. This highlights an unresolved gap: existing RAG frameworks lack a principled mechanism to integrate semantic complexity into both planning and verification.

To fill this gap, we propose PAR-RAG, a novel reasoning framework that instantiates a three-phase Plan-then-Act-and-Review paradigm inspired by the PDCA cycle \cite{r:14}, which emphasizes iterative planning, execution, and evaluation to continuously improve process outcomes. PAR-RAG embeds semantic complexity as a unifying principle that organizes the reasoning process. In planning, exemplar cases matched on inferred complexity condition plan generation so that decomposition granularity aligns with task difficulty—this reduces early step misspecification and stabilizes the reasoning trajectory. In acting, the system executes the plan via retrieve-then-read steps to collect and integrate evidence. In reviewing, a dual verification module adapts to complexity: low-complexity queries are verified primarily by accuracy, while high-complexity queries require multi-evidence factual consistency and cross-source corroboration, thereby improving credibility.

Through this integration of semantic complexity into each stage of reasoning, PAR-RAG provides a unified and cognitively grounded solution to the long-standing trade-off between reasoning stability and factual reliability. While belonging to the family of planning-guided methods, PAR-RAG fundamentally advances the paradigm by introducing adaptive planning and dual verification driven by semantic complexity. This design establishes a clear Plan-then-Act-and-Review reasoning cycle, enabling dynamic adjustment of reasoning granularity and verification strength according to task difficulty. 
\begin{figure}[t]
\centering
\includegraphics[width=1\columnwidth]{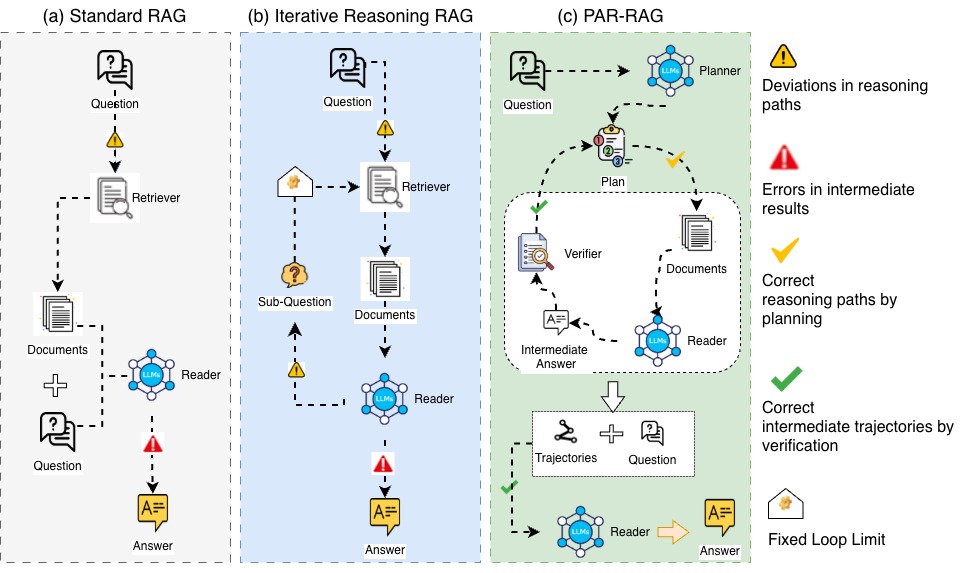} 
\caption{Comparison of the proposed PAR-RAG (c) with Standard RAG (a), Iterative Reasoning RAG (b).}
\label{figure1}
\end{figure}

As shown in Figure~\ref{figure1}, compared with prior paradigms—Standard RAG (including structure-aware variants), which relies on static retrieval–generation pipelines, and Iterative Reasoning RAG (including reasoning-enhanced and reflection-based approaches), which often lacks global coherence and incurs reasoning drift—PAR-RAG achieves a unified and adaptive multi-hop reasoning framework. It effectively balances local factual precision with global reasoning stability, forming a self-regulating process that continuously refines its reasoning outcomes and enhances interpretability.

Our main contributions are summarized as follows:
\begin{itemize}
    \item We propose PAR-RAG, which operationalizes a Plan-then-Act-and-Review mode to address reasoning drift and factual inconsistency in MHQA, explicitly separating planning, acting, and reviewing as complementary stages.
    \item We propose two transferable principles—(i) Complexity-Aware Reasoning, aligning strategies with task difficulty, and (ii) Dual Verification, enforcing factual reliability at multiple levels. These principles extend beyond QA to inform the broader design of trustworthy reasoning systems.
    \item Extensive experiments on various benchmarks show consistent improvements, with the most substantial gains in high-complexity datasets. Ablation studies confirm the complementary effects of planning and verification, highlighting the scalability of our design to tasks where correctness outweighs efficiency.
\end{itemize}

\section{Related Work}
Recent advances in retrieval-augmented generation (RAG) have produced a rich taxonomy of methods aimed at improving multi-step reasoning. We organize prior work into four categories—structure-aware, reasoning-enhanced, reflection-based, and planning-guided—and summarize their strengths and limitations with respect to two preferences: (i) constructing stable multi-step reasoning trajectories and (ii) providing systematic factual verification. We emphasize a recurring shortfall across these families: the lack of a principled mechanism that integrates semantic complexity to jointly guide plan generation and verification.
\begin{table*}
\begin{center}
\caption{A comprehensive comparison of capabilities across multiple RAG categories. \checkmark=Full Support, $\times$=None, $\bigtriangleup$=Partial.}
\label{table1}
\begin{tabular}{l c c c c c}
\hline
Method & Planning & Step Verification & Complexity Adaptation & Path Stability\\
\hline
Standard RAG &
$\times$ &
$\times$ &
$\times$ &
$\times$ \\
Structure-Aware RAG &
$\times$ &
$\times$ &
$\times$ &
$\bigtriangleup$ \\
Reasoning-Enhanced RAG &
$\bigtriangleup$ &
$\bigtriangleup$ &
$\times$ &
$\bigtriangleup$ \\
Reflection-Based RAG &
$\bigtriangleup$ &
\checkmark &
$\times$ &
$\bigtriangleup$ \\
Planning-Guided RAG &
\checkmark &
$\times$ &
$\times$ &
\checkmark \\
PAR-RAG &
\checkmark &
\checkmark &
\checkmark &
\checkmark \\
\bottomrule
\end{tabular}
\end{center}
\end{table*}
\subsection{Structure-Aware RAG}
Structure-aware methods improve retrieval granularity and contextual representation by transforming raw text into richer structural forms. Representative approaches include recursive hierarchical indexing \cite{r:18, r:23}, entity-relation graphs for multi-level retrieval \cite{r:19}, adaptive document-structure selection \cite{r:22}, and long-term memory pattern modeling \cite{r:20}. More recent work extends this direction to layout- and modality-aware graph modeling (e.g., \cite{r:66}), demonstrating substantial gains on tasks that demand fine-grained structural understanding. However, structure-aware systems primarily focus on input representation and retrieval quality; they typically do not implement explicit plan generation nor stage-wise verification mechanisms. Consequently, while structure can reduce noise in retrieved evidence, it does not by itself provide a mechanism to construct or verify multi-step reasoning trajectories—particularly when task complexity calls for adaptive plan granularity and variable verification strength.

\subsection{Reasoning-Enhanced RAG}
Reasoning-enhanced frameworks tightly couple LLM reasoning with retrieval, often via iterative retrieve-and-reason loops or multi-agent orchestration. Examples include Chain-of-Thought \cite{r:25} augmented retrieval \cite{r:24, r:54}, iterative retrieval-generation hybrids \cite{r:26}, and recent multi-agent orchestration systems (e.g., \cite{r:69, r:70}) that coordinate specialized agents (planners, extractors, verifiers) to decompose and solve complex queries. These methods improve interpretability and can mitigate some hallucinations through coordinated reasoning. Nonetheless, they frequently lack a global planning scaffold that enforces a stable trajectory across all steps; moreover, although some incorporate localized verification, they do not normally adapt verification policies according to instance difficulty in a principled way.

\subsection{Reflection-Based RAG}
Reflection-based approaches focus on introspective evaluation and post-hoc correction of generated outputs. Systems such as Self-RAG \cite{r:38}, Reflexion \cite{r:56}, and CRAG \cite{r:39} apply self-critique, re-generation, or targeted edits to reduce factual errors. Complementary evaluation efforts (e.g., \cite{r:64, r:65}) have developed fine-grained metrics to reveal persistent intermediate inconsistencies even when end-task accuracy appears acceptable. Reflection methods can improve answer factuality, but they commonly assume that the model can reliably self-diagnose—an assumption undermined by model hallucination and miscalibrated confidence \cite{r:36, r:37}. Crucially, most reflection pipelines apply the same verification strategy across instances, rather than dynamically scaling checks by semantic complexity. 

\subsection{Planning-Guided RAG}
Planning-guided methods explicitly produce intermediate plans prior to retrieval and generation. Classic examples include tree-search or agent-based planners \cite{r:28, r:29, r:58}, and recent work \cite{r:67} emphasizes structured, DAG-style plans to coordinate subqueries and evidence collection. These approaches are effective at reducing ad hoc reasoning fragmentation and improving alignment between subqueries and external evidence. Yet, many planning-guided systems rely on static exemplars or fixed planning heuristics, and few integrate an adaptive verification stage that modulates verification intensity according to task difficulty.

\subsection{Positioning PAR-RAG}

In sum, while each category contributes valuable techniques—structure for richer retrieval, reasoning-enhanced coordination for interpretability, reflection for post-hoc correction, and planning for structured decomposition—none provides a principled, end-to-end mechanism that (i) uses semantic complexity to condition exemplar selection and plan granularity, and (ii) dynamically adapts verification (coarse vs. fine) to instance difficulty. PAR-RAG addresses this gap by threading semantic complexity across Plan-then-Act-and-Review: complexity-aware exemplar selection anchors plan generation, the acting stage executes the structured plan to collect and integrate evidence, and dual verification adapts its focus from accuracy to multi-evidence consistency as complexity grows. As summarized in Table \ref{table1}, this combination yields both trajectory stability and systematic verification, distinguishing PAR-RAG from prior RAG variants.
\section{Methodology}
\begin{figure*}[t]
\centering
\includegraphics[width=1\columnwidth]{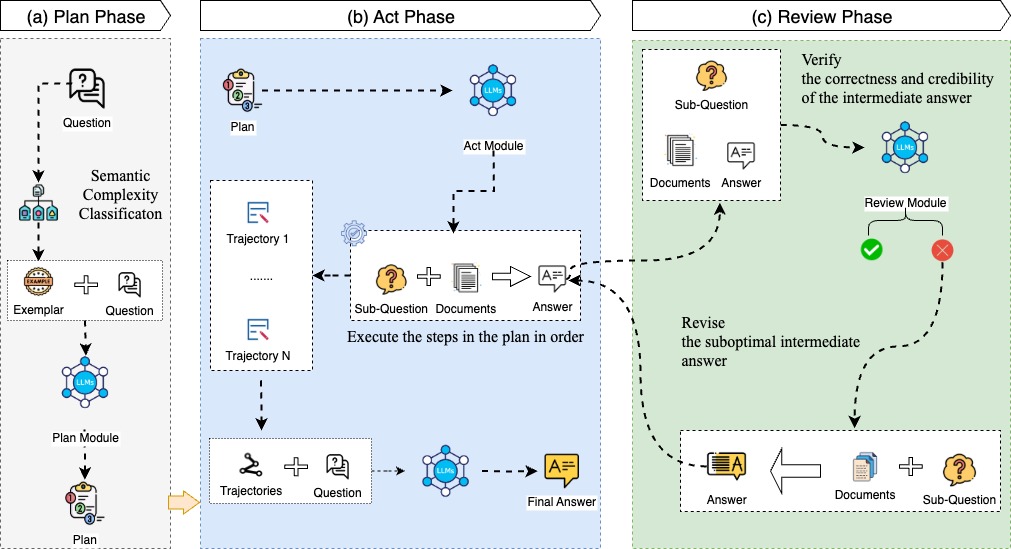} 
\caption{Overview of the PAR-RAG workflow, which follows a Plan-then-Act-and-Review cycle: planning generates exemplar-aligned reasoning steps, acting executes them to produce answers, and reviewing verifies the intermediate results for accuracy and consistency.}
\label{figure2}
\end{figure*}

In this section, we present the design of PAR-RAG, detailing how it operationalizes the proposed reasoning paradigm. As illustrated in Figure~\ref{figure2}, each stage in PAR-RAG corresponds to a cognitively inspired function: planning stabilizes the reasoning trajectory, acting executes retrieval and evidence integration, and reviewing enforces factual verification through dual mechanisms. Together, these stages form a closed-loop reasoning process that balances trajectory stability and factual reliability.

This design not only enhances empirical accuracy but also embodies two generalizable design principles: Complexity-Aware Reasoning, which adapts reasoning strategies to task difficulty, and Dual Verification, which enforces multi-level factual control throughout the reasoning process.

The remainder of this section is organized as follows. We first define the problem in Section~\ref{problem_formulation}, then present the overall workflow of PAR-RAG in Section~\ref{workflow}. Sections~\ref{semantic_complexity_aware_plan_generation} and~\ref{dual_verification} elaborate on the semantic complexity-aware plan generation and the dual verification mechanisms, respectively. (\textit{Note:} For clarity, we summarize the key symbols and variables in a notation table provided in Appendix~\ref{notation_table}).

\subsection{Problem Formulation}\label{problem_formulation}
Let a multi-hop question be denoted as a pair
$Q = \{q, (q_1, q_2, \dots, q_m)\}$,
where $q$ is the original query and $(q_1, q_2, \dots, q_m)$ represents its decomposition into $m$ sub-questions. Let the document collection be
$D = (d_1, d_2, \dots, d_k)$.
The objective of a plan-driven RAG framework is to construct a reasoning plan
$P = (p_1, p_2, \dots, p_m)$,
where each step $p_i$ corresponds to a sub-question $q_i$ and specifies how evidence is to be retrieved and processed. For each step, the system retrieves relevant evidence: $d_i = f_{\text{Retrieve}}(q_i)$, and generates an intermediate answer:
$a_i = f_{\text{Read}}(q_i, d_i)$,
where $f_{\text{Retrieve}}$ is the retrieval function (e.g., a dense retriever or an LLM-based retriever) and $f_{\text{Read}}$ is the reader function (typically an LLM).

Then, the intermediate data $(q_i, a_i, d_i)$ are consolidated into a trajectory record:
\[
t_i = f_{\text{GenerateTrajectory}}(q_i, a_i, d_i).
\]
Finally, the final answer is inferred from the complete reasoning trajectories:
\[
A_{\text{final}} = f_{\text{Read}}(q, t_1, t_2, \dots, t_m).
\]

Distinct from prior plan-driven RAG, PAR-RAG explicitly incorporates semantic complexity into both planning and verification. In planning, semantic complexity conditions exemplar selection to enhance the stability of plan generation for producing $P$; in reviewing, a verification function dynamically adapts its strength—emphasizing accuracy checks for low-complexity queries and multi-evidence factual consistency for high-complexity ones. This integration ensures trajectory stability and trustworthy verification across varying task difficulties.

\subsection{Workflow of PAR-RAG}\label{workflow}
Based on the proposed model architecture and problem formulation, the workflow of PAR-RAG unfolds through the following stages, as outlined in Algorithm~\ref{alg:algorithm}:
\begin{itemize}
    \item PAR-RAG first estimates the semantic complexity of a given multi-hop question using a BERT-based classifier. Conditioned on this estimate, it retrieves in-context exemplars with comparable complexity to anchor plan generation. Leveraging LLM reasoning capabilities, the system then produces a structured plan $P=(p_1,\dots,p_m)$, where each step $p_i=(\text{Thought}_i, Q_i)$ contains an interpretable reasoning trace and a decomposed sub-question $Q_i$. This ensures that each step is both executable and aligned with the difficulty of the input.
    \item Guided by the generated plan, each sub-question $Q_i$ is used for dense retrieval of relevant documents, $D_i=\text{DenseRetrieve}(Q_i)$. The system generates an intermediate answer $A_i=\text{Read}(Q_i,D_i)$ and extracts cited evidence $E_i=\text{ExtractCitation}(D_i)$ through citation filtering, thereby maintaining contextual relevance and constraining evidence usage.
    \item To enhance reliability, PAR-RAG applies dual verification to each $A_i$. Depending on the estimated semantic complexity, the system emphasizes accuracy for low-complexity queries and factual consistency (via multi-evidence corroboration) for high-complexity ones. If verification fails, the system performs sparse re-retrieval to broaden coverage for $Q_i$, and produces a revised answer $A_i=\text{Read}(Q_i,D_i)$. This adaptive mechanism jointly reduces reasoning drift and factual errors.
    \item Each verified step is appended to a reasoning trajectory $T_i=\text{GenerateTrajectory}(Q_i,A_i,E_i)$, forming the trajectory chain $T=(T_1,\dots,T_m)$. To preserve coherence, the next sub-question $Q_{i+1}$ may be refined based on $A_i$, mitigating error propagation across steps and maintaining global trajectory stability.
    \item Once all steps are completed, PAR-RAG uses the initial question $Q$ and the full trajectory $T$ as context to generate the final answer $A_{\text{final}}=\text{GenerateAnswer}(Q,T)$. The final output thus reflects both trajectory stability and complexity-adaptive verification.
\end{itemize}

This workflow instantiates the Plan-then-Act-and-Review paradigm: semantic complexity guides planning and verification, execution operationalizes retrieval and reading, and trajectory construction ensures stable, interpretable multi-hop reasoning.

\begin{algorithm}
\caption{Credible Plan-Driven RAG}
\label{alg:algorithm}
\textbf{Input}: Question $Q$\\
\textbf{Parameter}: Confidence Threshold $C_t$, Prompt Template $PT$\\
\textbf{Output}: $A_{final}$
\begin{algorithmic}[1] 
\State Let $i=1$.
\State Let $H_{complexity}=PredictHopCount(Q)$. \Comment{Predict the number of reasoning steps required for question $Q$, denoted as $H_{complexity}$.}
\State Let $E_{exemplar}=FindExemplarCases(Q, H_{complexity})$. \Comment{Identify exemplar cases for question $Q$ based on the predicted reasoning steps $H_{complexity}$, denoted as $E_{exemplar}$.}
\State Let $P=GeneratePlan(E_{exemplar}, Q, PT)$. \Comment{Generate a reasoning plan $P$ for question $Q$ using contextual references $E_{exemplar}$ and prompt template $PT$.}
\Repeat
\State $P_i \gets P[i]$ \Comment{Retrieve the $i$-th step of the plan, denoted as $P_i$.}
\State $Q_i \gets ExtractSubQuestion(P_i)$ \Comment{Extract the sub-question for the current step, denoted as $Q_i$.}
\State $D_i=DenseRetrieve(Q_i)$
\State $A_i=Read(Q_i, D_i)$ \Comment{Generate the intermediate answer $A_i$ based on the sub-question $Q_i$ and retrieved documents $D_i$.}
\State $E_i=ExtractCitation(A_i, D_i)$ \Comment{Extract citations, denoted as $E_i$, from the retrieved documents $D_i$ supporting the intermediate answer $A_i$.}
\State $C_i=EvaluateConfidence(Q_i, A_i, E_i)$
\If {$C_i < C_t$}
\State $D_i=SparseRetrieve(Q_i)$
\State $A_i=Read(Q_i, D_i)$
\State $E_i=ExtractCitation(A_i, D_i)$
\EndIf
\State $Q_{next} \gets P[{i+1}]$
\State $Q_{next}=RefineNextQuestion(A_i, Q_{next})$ \Comment{Refine the next sub-question $Q_{next}$ using the intermediate answer $A_i$.}
\State $T = Concatenate(T, GenerateTrajectory(Q_i, A_i, E_i))$ \Comment{Append the intermediate reasoning trajectory, generated from $Q_i$, $A_i$, and $E_i$, to the cumulative trajectory $T$.}
\State $i=i+1$
\Until{$i >= length(P)$}
\State $A_{final}=GenerateAnswer(Q, T)$
\State \textbf{return} $A_{final}$
\end{algorithmic}
\end{algorithm}

\subsection{Semantic Complexity-Aware Plan Generation}\label{semantic_complexity_aware_plan_generation}
Existing planning approaches for multi-hop question answering typically rely on either (i) zero-shot direct generation or (ii) prompt learning with fixed exemplars. However, both often overlook semantic complexity alignment between target questions and exemplars. Recent work shows that exemplar choice that accounts for demonstration complexity or reasoning structure substantially improves complex reasoning performance, while poorly matched or ordered exemplars can degrade or destabilize in-context reasoning \cite{r:50, r:51, r:59, r:60}.

To overcome this, we propose a semantic complexity-aware planning mechanism that estimates question complexity and retrieves exemplars of similar complexity as contextual references, thereby enhancing plan stability.

Since the semantic complexity of multi-hop questions is difficult to measure directly, we approximate it using the number of reasoning hops and train a classifier to predict hop counts. However, entropy-based measures alone are insufficient for capturing semantic complexity. Recent studies emphasize the necessity of combining handcrafted linguistic features with embedding-based representations for a more accurate assessment of semantic or lexical complexity. For instance, \cite{r:61} demonstrated that integrating deep contextual encodings (e.g., BERT) with more than twenty handcrafted features yields robust performance in lexical and sentence-level complexity prediction. Similarly, \cite{r:62} combined embeddings with lexical and syntactic indicators to improve lexical complexity prediction. Moreover, \cite{r:63} systematically examined instance-level complexity metrics across classification tasks and concluded that relying on a single indicator is inadequate, instead advocating for multi-feature integration encompassing lexical, syntactic, semantic, and embedding-based signals.  

Building on these findings, we design a multi-dimensional complexity modeling module that integrates linguistic and reasoning-related features with embedding representations, thereby providing a more comprehensive and robust latent representation of semantic complexity.

Specifically, we incorporate:
\begin{itemize}
    \item Syntactic depth: the maximum dependency tree depth, capturing the degree of structural composition.
    \item Semantic entropy: sentence-level word sense entropy, reflecting ambiguity and information density.
    \item Contextual embeddings: token-level BERT representations aggregated via pooling, encoding semantic richness.
    \item Number of named entities: the number of distinct entities mentioned, indicating potential cross-document reasoning demand.
    \item Sentence length: a surface-level proxy for structural and semantic load.
\end{itemize}

Sentence word sense entropy quantifies the semantic complexity of a sentence 
by measuring the uncertainty in the distribution of word senses for its constituent words. For a word $w_i$ with $k$ possible senses, and the probability of each sense denoted as $P(s_j \mid w_i)$,  
the word sense entropy is defined as:
\begin{equation}
SE(w_i) = -\sum_{j=1}^{k} P(s_j \mid w_i) \log P(s_j\mid w_i) .
\end{equation}

Extending this to the sentence level, the sentence word sense entropy aggregates the uncertainty 
across all words in a sentence consisting of $n$ words
\begin{equation}
MSE(\text{sentence}) = \frac{1}{n} \sum_{i=1}^{n} SE(w_i),
\end{equation}
where each $SE(w_i)$ represents the entropy of the sense distribution for word $w_i$.

A dependency tree is a syntactic structure derived from dependency parsing that models the syntactic relationships between words in a sentence. In this tree, each node corresponds to a word, and each edge represents a dependency relation between two words. The depth of a dependency tree is defined as the length of the path (measured by the number of edges) from the root node to the most distant leaf node—that is, a word with no children. In other words, the depth corresponds to the longest path from the root to any leaf node in the tree.

In summary, the semantic complexity of a given question \( q \) can be approximated as:
\begin{equation}\label{eq:crossatt}
\begin{aligned}
   \mathbf{x} = \big[ f_{\text{len}}(q), f_{\text{ent}}(q), f_{\text{dep}}(q), f_{\text{mse}}(q) \big],
\end{aligned}
\end{equation}
where each component represents a distinct linguistic or semantic attribute:
\begin{itemize}
    \item \( f_{\text{len}}(q) = |q| \) denotes the number of tokens after tokenization;
    \item \( f_{\text{ent}}(q) = |\mathrm{NER}(q)| \) denotes the number of named entities identified in \( q \);
    \item \( f_{\text{dep}}(q) = \max(\mathrm{depth}(T(q))) \) denotes the maximum depth of the dependency parse tree \( T(q) \);
    \item \( f_{\text{mse}}(q) = \mathrm{MSE}(q) \) denotes the semantic entropy of the question, reflecting contextual ambiguity.
\end{itemize}

The question $q$ is further encoded using a pretrained BERT encoder:
\begin{equation}
\mathbf{h}_q = \mathrm{BERT}(q)
\end{equation}
and the final representation is obtained by concatenating semantic embeddings and linguistic features:
\begin{equation}
\mathbf{z} = [\mathbf{h}_q ; \mathbf{x}].
\end{equation}
A pretrained classifier is then applied to $\mathbf{z}$ to predict the hop count of the target instance:
\begin{equation}
\mathrm{HopCount}(q) = \mathrm{f_{cls}}(\mathbf{z}),
\end{equation}
where HopCount(q) $\in$ \{1,2,3,4\} and $f_{cls}$ denotes a lightweight MLP-based classification head.
The predicted hop count serves as a retrieval condition to select exemplars of comparable complexity from the repository, thereby enhancing the stability and robustness of plan generation.

This design achieves two objectives. First, it bridges latent reasoning demand and observable features by learning a predictive proxy for hop count. Second, it ensures granularity alignment: exemplars with comparable predicted complexity better match the decomposition requirements of the input, reducing variance in plan generation and mitigating both oversimplification and unnecessary expansion.

Empirical results demonstrate that complexity-aware retrieval outperforms zero-shot and semantic similarity-based baselines in final QA accuracy.
\subsection{Dual Confidence Verification Mechanism}\label{dual_verification}
Existing RAG methods typically evaluate performance solely based on answer accuracy, often neglecting the factual consistency of generated results, specifically whether answers are genuinely supported by the retrieved evidence. To address this limitation, this study proposes a comprehensive evaluation mechanism that integrates both accuracy and factual consistency. This mechanism quantifies answer quality at each intermediate step in multi-hop question answering, enabling dynamic decisions regarding further retrieval or answer revision, thus improving the reliability of the final output. 

Specifically, given the question $q$, the corresponding answer $a_q$, and the contextual references $d_q$, we adopt the LLM-as-a-judge approach \cite{r:53}, where the LLM scores the accuracy of the generated answer:
\begin{equation}
    A = LLM(q, a_q, d_q).
\end{equation}
The output is a normalized score within the range [0, 1]. 

The factual consistency of the answer is evaluated using AttrScore \cite{r:55}, which assesses whether the generated statement is fully supported by the cited references. The three values returned by AttrScore—[Contradictory, Extrapolatory, Attributable]—are mapped to corresponding numerical scores [0, 0.5, 1], respectively, as follows:
\begin{equation}
    F = AttrScore(q, a_q, d_q) .
\end{equation}

The final confidence score is defined as the weighted combination of two factors: accuracy and credibility. To better adapt verification to question difficulty, we redefine the weight \(\alpha\) as a function of the predicted hop count \(\hat{h}\). This function follows a logistic mapping, enabling a smooth transition from accuracy-dominant weighting (for small \(\hat{h}\)) to credibility-dominant weighting (for larger \(\hat{h}\)):
\begin{equation}
\alpha(\hat{h}) \;=\; \frac{1}{1 + \exp\!\big(\gamma(\hat{h} - h_0)\big)},
\qquad \gamma>0,\; h_0>0,
\end{equation}
where $\gamma$ controls the steepness of the transition, and $h_0$ denotes the inflection point corresponding to the hop count at which accuracy and credibility are equally weighted.

The adaptive confidence score \(C_q\) is then computed as:
\begin{equation}
\mathrm{C_{q}} \;=\; A *{\alpha(\hat{h})} + F *({1-\alpha(\hat{h})}).
\end{equation}

Finally, the confidence score is compared against a threshold \(C_t\) to decide whether to accept the current answer or trigger further retrieval:
\begin{equation}
\text{Confidence}(a_q) = \begin{cases} 
1 & \text{if } C_{q} \geq C_t \\ 
0 & \text{if } C_{q} < C_t
\end{cases}
\end{equation}.
If the confidence equals 1, the answer is deemed both accurate and credible; otherwise, PAR-RAG performs an additional retrieval and revises the response for the current sub-question to improve intermediate results.

In practice, based on the results of the ablation study presented in Appendix~\ref{ablation_study_for_hyperparameters}, we set \(C_t = 0.65\), \(h_0 = 3.5\), and \(\gamma = 1.5\). This configuration yields a smooth yet decisive transition: accuracy is prioritized for single-hop questions, whereas factual consistency becomes dominant for higher-hop questions. Moreover, it stabilizes the weighting mechanism and mitigates over-sensitivity to errors in hop count prediction.
\section{Experiments}
\subsection{Benchmark Datasets}
To thoroughly evaluate the performance of PAR-RAG on multi-hop question answering, we selected three multi-hop QA datasets commonly used by other RAG methods:
\begin{itemize}
\item 2WikiMultiHopQA \cite{r:40}. It is constructed from Wikipedia content and knowledge graph paths, and is specifically designed to evaluate models’ capability in performing primarily two-hop reasoning tasks, while also providing support for more general multi-hop reasoning. 
\item HotpotQA\cite{r:4}. It challenges models with questions that bridge multiple Wikipedia articles, necessitating the integration of information from different sources. 
\item MuSiQue\cite{r:41}. It introduces complexity by combining multiple single-hop queries into multi-hop questions, requiring models to navigate through 2-4 logical steps.
\end{itemize}
Additionally, to validate whether PAR-RAG performs consistently on single-hop question-answering tasks, we selected the following single-hop dataset as evaluation source:
\begin{itemize}
    \item TriviaQA\cite{r:44}. It draws upon trivia questions from diverse online quiz platforms, offering a challenging testbed for open-domain question-answering systems.
\end{itemize}
We randomly selected 500 examples from each dataset to ensure comparable evaluation scale.
\subsection{Baseline Methods}
To comprehensively assess performance, we compared our method with representative baselines covering key RAG paradigms: 
\begin{itemize}
    \item Standard RAG: Standard RAG \cite{r:1} directly retrieves relevant evidence based on the question, selecting the top-k retrieved results as context data. These results, along with the question, are passed to the LLM through a prompt to generate the answer. 
    \item Structure-Aware RAG
    \begin{itemize}
        \item RAPTOR: RAPTOR \cite{r:18} recursively organizes document fragments into hierarchical summaries, constructing a tree where each node represents the summary of its child nodes. During retrieval, RAPTOR evaluates the relevance of each node in the tree and identifies the most relevant nodes, with the number of nodes limited by the top-k criterion. 
        \item HippoRAG w/ IRCoT: HippoRAG \cite{r:20} simulates the long-term memory mechanism of the human brain by extracting entities and their relationships from documents to build a knowledge graph. It establishes an index and efficiently retrieves relevant data using the Personalized PageRank (PPR) algorithm. It can be combined with IRCoT \cite{r:24} to enhance QA capabilities, offering a unique advantage in complex reasoning tasks. 
    \end{itemize}
    \item Reasoning-Enhanced RAG
      \begin{itemize}
        \item ReAct \cite{r:54} is an enhanced generation model framework that integrates reasoning and acting by alternating between reasoning steps and action steps, enabling the model to dynamically plan and execute in complex tasks. Compared to traditional methods, ReAct demonstrates higher flexibility and accuracy in scenarios such as question answering and task planning, especially excelling in tasks that require multi-step reasoning and environmental interaction.
        \item IRCoT \cite{r:24} is a framework that enhances language model performance through iterative reasoning and chain-of-thought prompting. By performing multiple rounds of iterative reasoning combined with chain-of-thought prompts, it progressively improves the model’s ability to answer complex questions.
      \end{itemize}
       \item Reflection-Based RAG
      \begin{itemize}
          \item Self-RAG \cite{r:38} introduces a reflection-based mechanism in retrieval-augmented generation, where the model generates self-critique tokens to assess the sufficiency and consistency of evidence. This reflective loop guides additional retrieval or revision, thereby improving reliability in multi-hop and complex question answering.
      \end{itemize}
      \item Planning-Oriented RAG
      \begin{itemize}
          \item Self-Ask \cite{r:58} addresses complex question answering by iteratively generating and answering sub-questions via external knowledge sources, achieving strong performance on benchmarks like HotpotQA and MuSiQue. Its effectiveness, however, is contingent on the quality of sub-question decomposition and the reliability of retrieved information.
      \end{itemize}
\end{itemize}

\subsection{Evaluation Metrics}
We used the following metrics to evaluate the performance of the methods:  
\begin{itemize}
    \item Exact Match (EM) \cite{r:46}: It measures whether the predicted answer precisely matches the expected result, focusing on exact matches. 
    \item Accuracy (Acc): This metric evaluates the semantic consistency between the generated answer and the reference answer using LLMs \cite{r:57}, thereby measuring the accuracy of the output. Compared to traditional surface-level matching methods (such as Exact Match or F1), this approach focuses more on semantic alignment, offering a more faithful reflection of answer quality in complex question answering tasks.
\end{itemize}

\subsection{Model Training}
In this section, we describe the training procedure and results of the semantic complexity classifier.  

\subsubsection{Training Data Construction}
To construct the training and validation sets for the semantic complexity classifier, we sampled multi-hop QA instances with hop counts ranging from 1 to 4. For the training set, we selected 1,171 instances per hop level from the training splits of MuSiQue and TriviaQA, ensuring balanced class distribution given the available instances, resulting in a total of 4,684 training examples. For validation, we adopted the same sampling strategy on the validation splits of MuSiQue and SQuAD, obtaining 850 instances for each hop level (1-4), thereby forming a validation set of 3,400 examples.
\subsubsection{Model Construction Methods}
We compare three variants of the semantic complexity classifier:
\begin{itemize}
    \item \textbf{(a) BERT Classifier:} A baseline classifier using BERT representations guided by information entropy;
    \item \textbf{(b) VAE Model:} A variational autoencoder (VAE) that integrates information entropy with additional complexity dimensions;
    \item \textbf{(c) BERT Classifier with Multiple Features:} An extended BERT classifier that incorporates semantic distribution entropy along with other complementary dimensions.
\end{itemize}

\subsubsection{Training Results}\label{training_results}
\begin{table*}[t]
\centering
\caption{Performance comparison of different semantic complexity classifiers. 
The BERT-based classifier with multiple features achieves the highest accuracy (81.32\%), 
while the VAE model underperforms despite extensive training epochs.}
\label{table2}
\begin{tabular}{l|c|cc}
\toprule
\textbf{Model} & \textbf{Accuracy} & \multicolumn{2}{c}{\textbf{Best Parameters}} \\
\cmidrule(lr){3-4}
 &  & \textbf{learning rate} & \textbf{epoch} \\
\midrule
(a) BERT Classifier & 80.83\% & 3.10e-5 & 5 \\
(b) VAE Model & 67.52\% & 2.00e-5 & 300 \\
(c) BERT Classifier with Multiple Features & 81.32\% & 3.05e-5 & 5 \\
\bottomrule
\end{tabular}
\end{table*}

\begin{figure}[t]
\centering
\includegraphics[width=0.9\columnwidth]{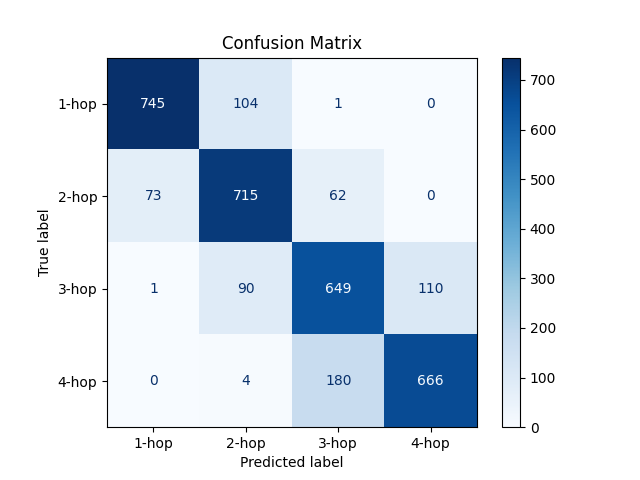} 
\caption{Confusion Matrix of the BERT classifier based on semantic entropy and other multi-dimensional features.}
\label{figure5}
\end{figure}

After multiple rounds of iterative training and cross-validation, as illustrated in Table \ref{table2}, the BERT-based classifier—enhanced with multi-dimensional features such as semantic entropy—achieved an accuracy of 81.32\% in predicting the semantic complexity (i.e., hop count) of multi-hop questions. However, as shown in Figure \ref{figure5}, a number of off-diagonal entries (e.g., 104 instances of 1-hop questions misclassified as 2-hop, and 90 instances of 3-hop predicted as 2-hop) reveal persistent confusion between adjacent hop levels. These results underscore the promise of semantic complexity-aware classification in guiding plan generation for multi-hop reasoning, while also highlighting the need for more fine-grained differentiation—an important direction for future research.

\subsection{Implementation Setup}
All experiments were conducted on a single MacBook Pro equipped with an Apple M4 chip 
(16-core CPU, 40-core GPU, 128GB unified memory). For large-scale training, we used a workstation 
with a single NVIDIA RTX 4090 GPU (24GB VRAM) and 128GB of system RAM. 

\subsubsection{Semantic Complexity Classifier} 
We adopted \texttt{bert-base-uncased} as the backbone to train a multi-dimensional BERT classifier that integrates semantic entropy and other complexity-related features, serving as the question complexity predictor within the PAR-RAG framework. The model was implemented in PyTorch~2.1.0 and optimized using the cross-entropy loss function.

\subsubsection{Exemplar Database} 
We constructed a vector database to store exemplar cases generated from both multi-hop and 
single-hop training datasets. Ten exemplar cases were assigned to each hop label, and by default, 
one exemplar case is retrieved to guide plan generation. 

\subsubsection{Dual Verification Model} 
For factual consistency checking, we used a fine-tuned AttrScore-FlanT5 (11B) model~\cite{r:55}. The Qwen 
model \footnote{\url{https://www.alibabacloud.com/help/en/model-studio/what-is-qwen-llm}} served as a judge to evaluate the accuracy of generated answers under instruction-following settings. 

\subsubsection{Settings of RAG Baselines} 
In all retrieval-augmented generation settings, we employed ColBERTv2 \cite{r:47} as the dense retriever, BM25S\footnote{\url{https://github.com/xhluca/bm25s}} as the sparse retriever, and GPT-4o as the LLM, with the Top-$K$ parameter fixed at 10 and the temperature parameter set to 0. To ensure a fair comparison, the vector database was constructed following the same procedure as in the standard RAG baseline, where the title and content of each document in the dataset were concatenated into a single text block for indexing.

\subsubsection{Prompts} 
The complete prompts used in our study are provided in the Appendix \ref{used_prompts}. 
\section{Results and Analyses} \label{results_and_analyses}
\subsection{Main Results}
To thoroughly assess the generalizability and robustness of PAR-RAG across diverse multi-hop question answering scenarios, we benchmarked its performance on three widely adopted datasets: 2WikiMultiHopQA (2Wiki), HotpotQA, and MuSiQue. As presented in Table \ref{table3}, PAR-RAG consistently outperforms state-of-the-art RAG baselines on nearly all metrics, including EM and overall answer accuracy. The only exception occurs on the EM score for 2WikiMultiHopQA, where PAR-RAG ties with the best-performing method (HippoRAG w/ IRCoT).

Quantitatively, PAR-RAG delivers substantial improvements over the second-best baseline across all datasets. Specifically, it achieves relative EM improvements of 16.39\% on 2Wiki, 5.17\% on HotpotQA, and 6.45\% on MuSiQue. In terms of overall accuracy, PAR-RAG consistently surpasses the second-best baseline with relative improvements of 2.63\%, 2.86\%, and 2.38\% on the three benchmarks, respectively. These gains are particularly notable on MuSiQue, where challenges such as complex question decomposition and distractor interference often hinder performance.

Moreover, PAR-RAG also demonstrates strong generalizability on single-hop reasoning: on TriviaQA, it achieves relative improvements of 9.09\% in EM and 3.08\% in accuracy compared with the strongest baseline. This confirms that PAR-RAG is not limited to multi-hop reasoning; rather, its core mechanism—adaptive plan generation guided by semantic complexity—remains effective for simpler single-hop tasks.
\begin{table*}[ht]
\begin{center}
\caption{Comparison of results across multiple datasets for various RAG methods. Bold and underline indicate the best and the second-best results.}
\label{table3} 
\begin{tabular}{*{10}{c}}
  \toprule
  \multirow{2}*{Name} & 
  \multicolumn{2}{c}{2Wiki} & 
  \multicolumn{2}{c}{HotpotQA} & 
  \multicolumn{2}{c}{MuSiQue} &
  \multicolumn{2}{c}{TriviaQA} \\
  \cmidrule(lr){2-3}\cmidrule(lr){4-5}
  \cmidrule(lr){4-5}
  \cmidrule(lr){6-7}
  \cmidrule(lr){8-9}
  & EM & Acc & EM & Acc & EM & Acc & EM & Acc\\
  \midrule
  Standard RAG & 0.37 & 0.38 & 0.45 & 0.52 & 0.08 & 0.08 & \underline{0.55} & 0.58\\
  RAPTOR & 0.16 & 0.22 & 0.26 & 0.35 & 0.06 & 0.12 & 0.34 & 0.44\\
  IRCoT & 0.49 & 0.61 & \underline{0.58} & \underline{0.70} & \underline{0.31} & 0.35 & 0.35 & 0.62\\
  HippoRAG w/ IRCoT & \textbf{0.71} & \underline{0.76} & 0.57 & \underline{0.70} & 0.30 & \underline{0.42} & 0.33 & \underline{0.65}\\
  ReAct & \underline{0.61} & 0.69 & 0.32 & 0.39 & 0.15 & 0.36 & 0.54 & 0.62 \\
  Self-Ask & 0.37 & 0.43 & 0.53 & 0.62 & 0.13 & 0.24 & 0.45 & 0.63\\
  Self-RAG & 0.02 & 0.06 & 0.15 & 0.21 & 0.02 & 0.02 & 0.15 & 0.26\\
  PAR-RAG & \textbf{0.71} & \textbf{0.78} & \textbf{0.61} & \textbf{0.72} & \textbf{0.33} & \textbf{0.43} & \textbf{0.60} & 
  \textbf{0.67} \\
  \midrule
  PAR-RAG (w/o Plan Module )& 0.51 & 0.53 & 0.5 & 0.63 & 0.16 & 0.23 & 0.51 & 0.54\\
  PAR-RAG (w/o Review Module) & 0.68 & 0.73 & 0.53 & 0.67 & 0.29 & 0.37 & 0.52 &0.59 \\
\bottomrule
\end{tabular}
\end{center}
\end{table*}
\subsection{Analysis}
\subsubsection{Ablation Study}
To assess the impact of the Plan Module and Review Module of PAR-RAG, we conducted ablation experiments involving two variants. The variant without the Plan Module treats the original question as a single-step plan, foregoing multi-step decomposition, while the variant without the Review Module disables the verification process, accepting intermediate answers without validation or revision. All other components remain unchanged.

As shown in Table \ref{table3}, both ablation variants significantly underperform compared to the full PAR-RAG model across all metrics and datasets. The variant lacking the Plan Module shows a more substantial decline, highlighting the essential function of global planning in multi-step reasoning. These results underscore the importance of the full PAR-RAG framework. 
\subsubsection{Effect of Sampling Strategies}
\begin{figure}[t]
\centering
\includegraphics[width=1\columnwidth]{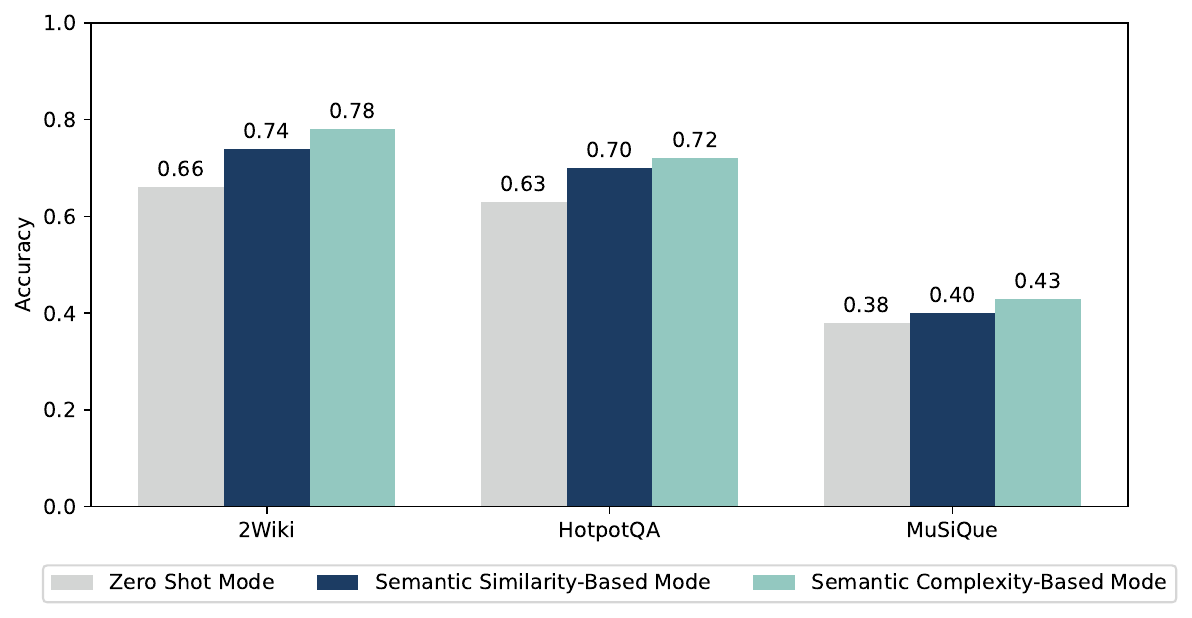} 
\caption{Accuracy Gains from Complexity-Aligned Exemplar Selection over Alternative Strategies.}
\label{figure3}
\end{figure}

This section examines how different example selection strategies influence the quality of multi-hop question plan generation. Accuracy (Acc) is used as the primary evaluation metric to ensure comparability. As shown in Figure \ref{figure3}, selecting examples whose semantic complexity closely matches that of the target question significantly enhances plan generation compared to zero-shot or semantic similarity-based strategies.

The observed gains can be attributed to the alignment of reasoning difficulty between exemplars and the target. Complexity-aligned exemplars provide not only semantic relevance but also coherent logical structures, thereby offering more effective guidance for capturing the latent reasoning processes required in multi-hop QA \cite{r:52}. By contrast, semantic similarity alone may retrieve superficially related exemplars that fail to reflect the necessary depth of reasoning. These results substantiate the effectiveness of the semantic complexity classifier in PAR-RAG: by ensuring complexity-matched exemplar selection, the model benefits from richer contextual support, which in turn enhances both the stability and accuracy of generated plans.
\subsubsection{Impact of Failed Semantic Complexity Prediction}
As detailed in Section~\ref{training_results}, the semantic complexity classifier achieves an accuracy of 81.32\%. This raises a critical question: how do misclassified hop counts affect the accuracy of the final answer? 

\begin{figure}
\centering
\includegraphics[width=1\columnwidth]{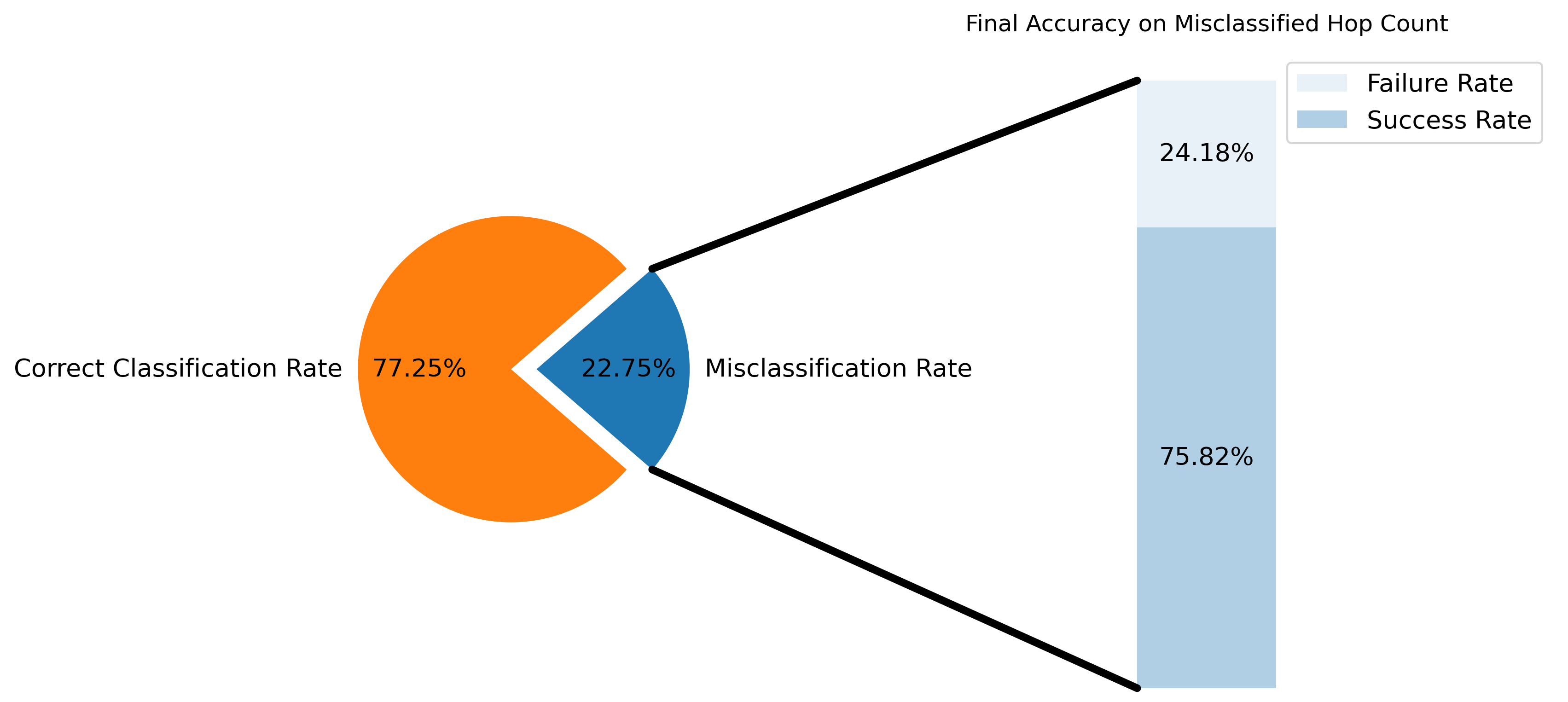} 
\caption{Impact of Misclassified Hop Counts on Final Answer Accuracy.}
\label{figure7}
\end{figure}

To investigate this, we conducted an experiment by collecting cases where hop count predictions were incorrect and recording the corresponding final answers. The evaluation dataset was constructed by randomly sampling 100 test cases from each benchmark dataset. As shown in Figure~\ref{figure7}, PAR-RAG achieves an overall accuracy of 77.25\% for predicting the hop count of the input question on these test datasets, and it demonstrates strong robustness: 75.82\% of the cases with incorrect hop count predictions still produced correct final answers. 

\begin{table}[t]
\begin{center}
\caption{Error distribution of hop count predictions and their corresponding effect on answer accuracy.}
\label{table4}
\begin{tabular}{l|l|l|l}
\hline
Hop Count & Incorrect Predictions & Success Rate & Failure Rate \\
\hline
Single Hop & 13 & 69.23\% & 30.77\% \\
Two Hops   & 47 & 80.85\% & 19.15\% \\
Three Hops & 15 & 60.00\% & 40.00\% \\
Four Hops  & 16 & 81.25\% & 18.75\% \\
\bottomrule
\end{tabular}
\end{center}
\end{table}

These results indicate that errors in hop count prediction do not necessarily propagate to the final answer, highlighting the resilience of PAR-RAG’s reasoning process. 

Moreover, Table~\ref{table4} shows that two-hop questions are especially prone to misclassification, likely because their semantic complexity is often underestimated and confused with adjacent hop levels, particularly single-hop questions. This highlights the need for more reliable methods of estimating semantic complexity to further improve prediction accuracy. Nevertheless, despite such misclassifications, PAR-RAG maintains high final-answer accuracy, demonstrating strong resilience to errors in hop count prediction.
\subsubsection{Case Study}
\begin{figure*}[t]
\centering
\includegraphics[width=\textwidth]{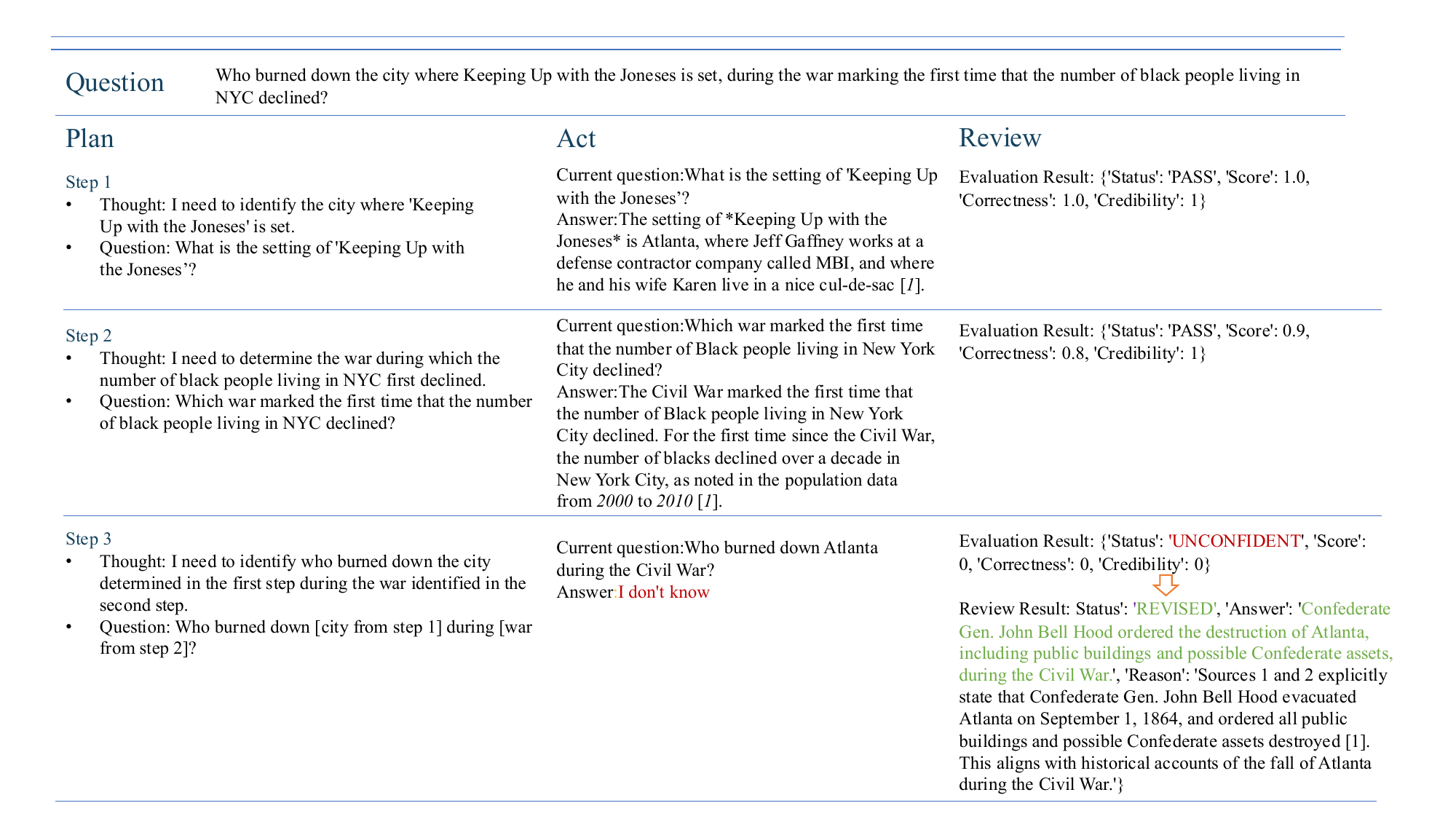} 
\caption{Illustration of PAR-RAG’s Planning and Review Mechanisms in multi-hop question answering.}
\label{figure4}
\end{figure*}

To illustrate the operational mechanism of PAR-RAG, we analyze a multi-hop question answering example presented in Figure \ref{figure4}. This example, sourced from a standard multi-hop dataset, exemplifies the model’s reasoning performance. From this analysis, we identify the following key insights:
\begin{itemize}
    \item Global planning guides PAR-RAG to produce logically coherent and well-structured multi-step reasoning plans for complex problems. By strictly adhering to these pre-generated steps, the model consistently derives accurate answers. 
    \item The review mechanism maintains the accuracy of reasoning outcomes. During plan execution, PAR-RAG employs a Review Module to continuously detect and assess intermediate results. Upon detecting potential errors (marked in yellow in the figure), the system corrects them using contextual information (marked in green), effectively halting error propagation and enhancing the accuracy and reliability of the final output.
\end{itemize}

\subsubsection{Error Analysis}
To gain deeper insights into PAR-RAG's performance, we performed a systematic error analysis on 30 failure cases drawn from four benchmark datasets. The findings, summarized in Table \ref{table5}, reveal several distinct categories of errors:
\begin{itemize}
    \item Planning Errors (27\%): These errors mainly arise from the LLM’s inadequate understanding of the semantic structure of complex questions, which produces reasoning plans misaligned with the intended query, thereby causing incorrect answers.
    \item Retrieval Errors (13\%): These errors generally stem from insufficient relevant information in the knowledge base, causing the retrieval stage to provide inadequate evidence to support effective reasoning.
    \item Reasoning and Generation Errors (60\% in total): Reasoning errors by the LLM constitute 43\%, making them the primary cause of failure. These arise when the retrieved information lacks explicit answers, and the LLM fails to integrate implicit clues for logical inference. Generation errors, accounting for 17\%, occur when the LLM produces incorrect answers during response generation despite retrieving relevant evidence. Collectively, these errors highlight the ongoing challenges faced by LLMs in semantic integration and response accuracy.
\end{itemize}

In summary, the analysis indicates that the primary limitations of current multi-hop question answering systems lie in their language understanding and cross-passage reasoning when handling complex queries. However, we acknowledge that the number of failure cases is limited, which may not fully reflect the true error landscape in question answering.
\begin{table}[t]
\begin{center}
\caption{Error types and their relative distribution based on 30 failure cases sampled from four benchmark datasets.}
\label{table5}
\begin{tabular}{l|l|l}
\hline
Error Category & Count & Percentage \\
\hline
Planning Error & 8 & 27\% \\
Retrieval Error & 4 & 13\% \\
Inference Error & 13 & 43\% \\
Generation Error & 5 & 17\% \\
\bottomrule
\end{tabular}
\end{center}
\end{table}
\subsubsection{Effect of Model Size}
As detailed in the section above, due to the substantial reasoning demands of knowledge-based tasks, systematically evaluating language models of varying parameter sizes within complex multi-hop contexts is essential to elucidate the relationship between model capacity and task effectiveness. 

To this end, we conducted comparative experiments involving several mainstream LLMs and pretrained language models (PLMs). Table \ref{table6} illustrates that models with smaller parameter sizes, such as GPT-4o-mini and Llama3-8B, perform notably worse on complex reasoning tasks compared to the more capable GPT-4o. These findings underscore that reasoning ability is a critical factor in addressing complex problems, reinforcing the necessity of utilizing models with robust inferential capacities in such contexts.
\begin{table}[t]
\begin{center}
\caption{Impact of model size on reasoning performance across multi-hop QA datasets.}
\label{table6}
\begin{tabular}{*{12}{c}}
  \toprule
  \multirow{2}*{Model} & 
  \multicolumn{2}{c}{2Wiki} & 
  \multicolumn{2}{c}{HotpotQA} & 
  \multicolumn{2}{c}{MuSiQue} \\
  \cmidrule(lr){2-3}\cmidrule(lr){4-5}
  \cmidrule(lr){4-5}
  \cmidrule(lr){6-7}
  & EM & Acc & EM & Acc & EM & Acc\\
  \midrule
  GPT-4o & \textbf{0.71} & \textbf{0.78} & \textbf{0.61} & \textbf{0.72} & \textbf{0.33} & \textbf{0.43} \\
  GPT-4o-mini & 0.51 & 0.59 & 0.45 & 0.55 & 0.26 & 0.31 \\
  Llama3-8B & 0.24 & 0.25 & 0.30 & 0.36 & 0.07 & 0.15 \\
\bottomrule
\end{tabular}
\end{center}
\end{table}
\subsubsection{Efficiency Analysis}
\begin{table*}[ht]
\begin{center}
\caption{Efficiency comparison of different RAG methods across four benchmark datasets, evaluated by average response time per query (RTPQ) and consumed tokens per query (CTPQ). The symbol $\downarrow$ indicates that lower values denote better performance. The best and second-best results are highlighted in \textbf{bold} and \underline{underline}, respectively.}
\label{table7}  
\begin{tabular}{*{10}{c}}
  \toprule
  \multirow{2}*{Name} & 
  \multicolumn{2}{c}{2Wiki} & 
  \multicolumn{2}{c}{HotpotQA} & 
  \multicolumn{2}{c}{MuSiQue} &
  \multicolumn{2}{c}{TriviaQA} \\
  \cmidrule(lr){2-3}\cmidrule(lr){4-5}
  \cmidrule(lr){4-5}
  \cmidrule(lr){6-7}
  \cmidrule(lr){8-9}
  & RTPQ$\downarrow$  & CTPQ$\downarrow$ & RTPQ$\downarrow$ & CTPQ$\downarrow$ & RTPQ$\downarrow$ & CTPQ$\downarrow$ & RTPQ$\downarrow$ & CTPQ$\downarrow$\\
  \midrule
  Standard RAG & \underline{1.73} & \textbf{922.35} & \underline{1.23} & \textbf{857.49} & \underline{1.61} & \textbf{906.10} & \textbf{0.84} & \textbf{954.11} \\
  RAPTOR & \textbf{1.07} & \underline{1070.29} & \textbf{0.98} & \underline{1041.97} & \textbf{1.14} & \underline{1018.79} & \underline{0.91} & \underline{1020.38} \\
  IRCoT & 4.40 & 1634.39 & 2.48 & 2248.59 & 3.48 & 2168.46 & 3.67 & 1902.11\\
  HippoRAG w/ IRCoT & 5.58 & 1795.44 & 4.89 & 2260.81 & 5.15 & 2239.91 & 6.44 & 1991.80\\
  ReAct & 23.66 & 8069.12 & 28.72 & 8759.90 & 22.43 & 12021.44 & 13.56 & 7676.08 \\
  Self-Ask & 8.96 & 11352.51 & 8.45 & 11219.00 & 9.18 & 13481.23 & 6.85 & 9728.54\\
  Self-RAG & 9.89 & 0.00 & 12.24 & 0.00 & 13.32 & 0.00 & 10.20 & 0.00 \\
  PAR-RAG & 17.18 & 4010.87 & 14.33 & 3501.06 & 21.17 & 4781.20 & 9.41 & 2345.91 \\
\bottomrule
\end{tabular}
\end{center}
\end{table*}

To measure the computational cost when handling the question answering tasks with various RAG methods, we employ the following evaluation metrics:
\begin{itemize}
    \item \textbf{RTPQ} (Response Time Per Query): This metric gauges the response speed of each method by measuring the average time (in seconds) it takes to respond to a query.
    \item \textbf{CTPQ }(Consumed Token Per Query): This metric measures the usage cost of each method by evaluating the average number of tokens consumed per query.
\end{itemize}
Table \ref{table7} reports the average response time per query (RTPQ) and consumed tokens per query (CTPQ) across different methods. As expected, PAR-RAG exhibits higher computational costs compared to standard RAG baselines, with response times ranging from 14s to 21s and token consumption between 3.5k and 4.7k. This overhead arises from the additional planning and dual-verification stages, which inevitably introduce latency and resource usage. (\textit{Note:} For Self-RAG, CTPQ is reported as 0 since it employs a self-trained model without external retrieval, and thus token consumption is not directly comparable to other baselines.)

However, such costs should be interpreted as a deliberate trade-off. Unlike purely efficiency-oriented methods, PAR-RAG prioritizes accuracy, factuality, and trustworthiness by enforcing a more controlled reasoning process. This makes the framework particularly valuable for high-stakes scenarios—such as medical consultation, education, and legal reasoning—where correctness is paramount and moderate latency is acceptable. In contrast, for real-time or latency-critical applications, efficiency remains a challenge. We therefore view the efficiency-trustworthiness trade-off as a key direction for future research, motivating the development of lightweight yet reliable verification strategies to reduce overhead while preserving the benefits of planning-driven reasoning.
\section{Discussion}
Our findings show that PAR-RAG consistently improves multi-hop QA by reducing reasoning drift and mitigating error propagation. Beyond empirical performance, the framework provides broader insights into complex reasoning within retrieval-augmented systems. Its semantic complexity-aware planning offers a structured scaffold that aligns with cognitive theories of forward planning and controlled inference, while its dual-verification module resembles meta-cognitive monitoring by continuously checking the factual soundness of intermediate steps. Together, these features suggest that PAR-RAG is not merely an engineering integration but an algorithmic instantiation of cognitively inspired reasoning mechanisms.

At the same time, several limitations must be acknowledged. First, the semantic complexity classifier achieves 81.32\% accuracy, which occasionally leads to suboptimal exemplar selection and unstable downstream reasoning. Future work could explore adaptive or self-supervised complexity estimation to reduce reliance on fixed classifiers. Second, the verification stage depends on LLM-as-a-judge, raising concerns about reproducibility, bias, and alignment with external evaluation standards. This limitation reflects a broader challenge in current RAG research: how to design reliable, interpretable verification mechanisms that go beyond LLM self-assessment. Third, PAR-RAG introduces additional computational overhead, as indicated by RTPQ and CTPQ metrics. While we view this as an efficiency-credibility trade-off—prioritizing correctness and reliability over speed—it underscores the need for adaptive computation strategies that can dynamically balance latency with trustworthiness depending on the application context.

Overall, PAR-RAG points to several future directions. Integrating richer complexity signals (e.g., semantic, syntactic, or domain-specific features), combining symbolic verification with neural methods to improve factual grounding, and designing adaptive execution strategies could further enhance both robustness and efficiency. These extensions would not only strengthen PAR-RAG itself but also contribute to the broader development of trustworthy and generalizable reasoning frameworks in retrieval-augmented generation.

\section{Conclusion and Future Direction}
This paper introduced PAR-RAG, a planning-driven and verifiable RAG framework that instantiates a three-phase Plan-then-Act-and-Review paradigm. By integrating semantic complexity as a unifying principle, PAR-RAG addresses two long-standing challenges in multi-hop QA—reasoning trajectory drift and insufficient factual verification—within a single coherent framework. In doing so, it reframes credibility as a first-class design objective for retrieval-augmented systems, beyond accuracy alone.

Our findings highlight two transferable design principles: (i) complexity-aware reasoning, which aligns decomposition granularity with task difficulty to stabilize reasoning trajectories, and (ii) dual verification, which enforces factual reliability at multiple levels. These principles not only advance the robustness of multi-hop QA but also provide generalizable insights for building trustworthy reasoning systems in high-stakes domains such as healthcare, law, and education.

Looking forward, future research may explore lightweight or adaptive verification mechanisms to improve scalability, richer linguistic and semantic signals for dynamic complexity modeling, and extensions of the paradigm to long-horizon conversational reasoning and multimodal settings. By advancing both theoretical grounding and practical robustness, PAR-RAG contributes to the broader agenda of developing credible, generalizable, and trustworthy RAG systems.

\section*{Declaration of competing interest} 
The authors declare that they have no known competing financial interests or personal relationships that could have appeared to influence the work reported in this paper.

\bibliography{references}

\clearpage
\appendix
\section{Ablation Study on Dual Verification Hyperparameters} \label{ablation_study_for_hyperparameters}
As described in Section~\ref{dual_verification}, the dual verification module involves several critical hyperparameters for computing the confidence score: the slope parameter $\gamma$, the inflection point of the hop count $h_0$, and the confidence threshold $C_t$.

To determine their optimal values, we conducted an ablation study by varying $\gamma \in \{0.1, 0.5, 1.0, 1.5, 2.0, 2.5\}$, \quad $h_0 \in \{1, 1.5, 2.0, 2.5, 3.0, 3.5\}$, and $C_t \in \{0.55, 0.65, 0.70, 0.75, 0.80, 0.85\}$. Each configuration was evaluated using a dataset built from intermediate reasoning steps during testing on both multi-hop and single-hop benchmarks.

\begin{figure}
\centering
\includegraphics[width=1\columnwidth]{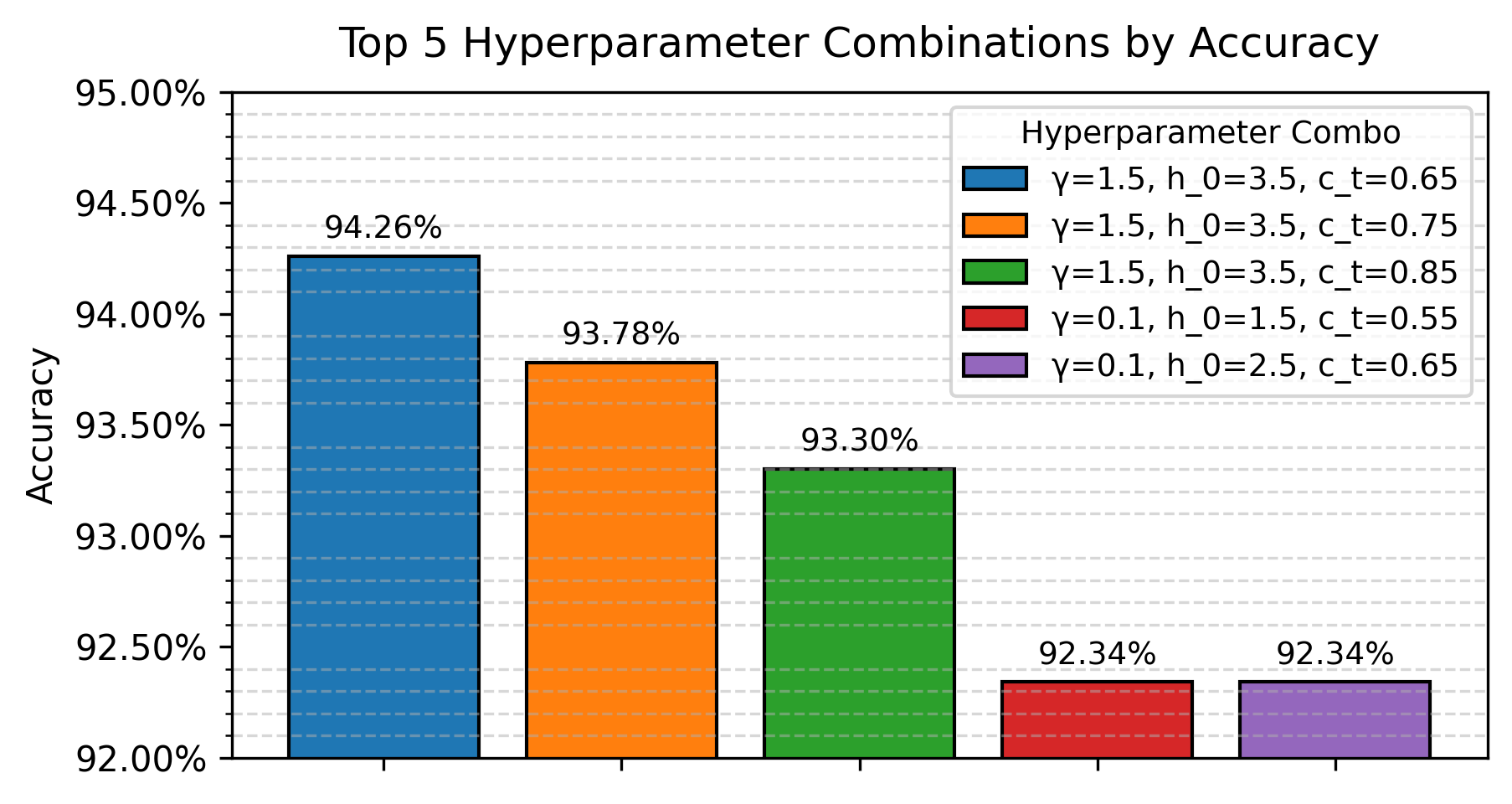} 
\caption{Performance Variation under Different $\gamma$, $h_{0}$, and $c_{T}$ Settings in Dual Verification. Here, $\gamma$ denotes the slope parameter, $h\_0$ is the inflection point of the hop count $h_0$, and $c\_t$ represents the confidence threshold $C_t$.}
\label{figure6}
\end{figure}

Accuracy was adopted as the evaluation metric, defined as the proportion of intermediate results for which the dual verification module reached the same judgment as human annotators. The best-performing configuration, shown in Figure~\ref{figure6}, corresponds to $\gamma=1.5$, $h_0=3.5$, and $C_t=0.65$, yielding an accuracy of 94.28\%.

These results not only confirm the robustness of the selected hyperparameter configuration but also highlight the effectiveness of employing an LLM-as-a-judge within the dual verification mechanism. In comparison to manual human evaluation, the module provides a reliable and fully automated means of validating intermediate reasoning steps, thereby enhancing both the stability and overall performance of the system.

\section{Notation Table}\label{notation_table}
\begin{table*}[t]
\centering
\caption{Summary of notation used in PAR-RAG.}
\label{table8}
\begin{tabular}{l|l}
\toprule
\textbf{Symbol} & \textbf{Definition / Notes} \\
\midrule
$q$ & Input question in natural language. \\
$z$ & Feature vector of $q$ (combination of syntactic depth, entropy, and embeddings; see Eq.~(1)--(5)). \\
$\text{HopCount}(q)$ & Predicted hop count of $q$; output of classifier, integer in $\{1,2,3,4\}$ (Eq.~(6)). \\
$\hat{h}$ & Predicted hop count label, used in planning and review modules. \\
$P$ & Reasoning plan: an ordered list of intermediate steps generated by the LLM. \\
$P_i$ & The $i$-th step of the reasoning plan $P$; $P_i \in P$. \\
$C_t$ & Confidence threshold controlling acceptance of factual consistency score (Eq.~(10)). \\
$h_0$ & Logistic midpoint for hop adjustment (Eq.~(9)). \\
$\gamma$ & Logistic growth rate controlling steepness (Eq.~(9)). \\
$\alpha(\hat{h})$ & Weighting factor for factual consistency; computed via logistic function (Eq.~(9)). \\
$\text{AttrScore}(q, a_q , d_q )$ & Attribution score of answer $a_q$ with respect to question $q$ and contextual references $d_q$ (Eq.~(8)). \\
$\text{Verify}(q, a_q, d_q)$ & Dual verification score combining accuracy and factual consistency (Eq.~(11)). \\
\bottomrule
\end{tabular}
\end{table*}

\section{Prompts of PAR-RAG} \label{used_prompts}
The prompts utilized in PAR-RAG are outlined as follows.
\begin{tcolorbox}[colback=gray!10,colframe=black,title=Plan Generation]
{You are a thoughtful expert who is very good at 
breaking down complex problems by developing a plan.
For each step in the plan, generate a sub-problem, 
each of which must be clear and independently executable, 
and all of which are logically coherent 
with the goal of solving the complex problem.
Do not use any external knowledge, assumptions, or information 
beyond what is explicitly stated in the context. 
\# Format instructions 
Use the following Strict JSON format, 
only choose an action from the list:[Retrieve, Answer]: 
[
  { 
    "Thought":"[your thought about the current step]"
    "Question":"[the question you generated for the current step]"
    "Action": "[the action you chose]"
  }
] }.
Don't output incomplete plan.
\end{tcolorbox}

\begin{tcolorbox}[colback=gray!10,colframe=black,title=Step Execution]
{You are an expert at inference and citation. 
You will read each source information carefully 
and cite evidence related to the current issue to answer the question.
When referencing information from a source, 
cite the appropriate source(s) using their corresponding numbers. 
Every answer should include at least one source citation. 
Only cite a source when you are explicitly referencing it.
If you don't know the answer, just return "I don't know".
Directly respond your answer, do not explain your answer 
or distract from irrelevant information in the source, 
and do not output anything that is not relevant to the question.
\# Examples begin
  Source 1:
  The sky is red in the evening and blue in the morning.
  Source 2:
  Water is wet when the sky is red.
  Source 3:
  Wolves and dogs belong to the species, Canis lupus.
  Query: When is water wet?
  Answer: In the evening. Water is wet when the sky is red[2], 
  which occurs in the evening [1].
\# Examples end.}
\end{tcolorbox}

\begin{tcolorbox}[colback=gray!10,colframe=black,title=Answer Verification]
{Given the following question, answer and context, 
evaluate the factual correctness on a scale from 0 to 1.
Question: {question}
Answer: {answer}
Context: {context}
Directly return the correctness score.
Don't explain yourself or output anything else.}
\end{tcolorbox}

\begin{tcolorbox}[colback=gray!10,colframe=black,title=Refine Next Question]
{You are good at analysis and inference.
Please only supplement the question with relevant 
and logical information based on the previous reasoning trajectories provided, 
without deviating from the original question intent.
Respond with the refined question only, 
do not explain yourself or output anything else.}
\end{tcolorbox}

\begin{tcolorbox}[colback=gray!10,colframe=black,title=Answer Generation]
{You serve as an intelligent assistant, 
adept at facilitating users through complex, 
multi-hop reasoning across multiple trajectories, 
each consisting of a document set including a reasoning thought, 
answer and relevant information. 
Your task is to generate the last answer for the question 
by referring to the trajectories.
If you don't know the answer, just return "I don't know".
Else, respond the final answer only, 
do not explain yourself or output anything else.
\# Example start
  Question: What is the distance in kilometers between Tokyo and Osaka, 
  rounded to the nearest whole number? 
  So the answer is: 550 
\# Example end.}
\end{tcolorbox}

\end{document}